# Cross-Covariance Modelling via DAGs with Hidden Variables

Jacob A. Wegelin
Thomas S. Richardson
University of Washington
Department of Statistics
Box 354322
Seattle WA 98195

## Abstract

DAG models with hidden variables present many difficulties that are absent when all nodes are observed. In particular, fully observed DAG models are identified and correspond to well-defined sets of distributions, whereas this is not true if nodes are unobserved. In this paper we characterize exactly the set of distributions given by a class of Gaussian models with one-dimensional latent variables. These models relate two blocks of observed variables, modeling only the cross-covariance matrix. We describe the relation of this model to the singular value decomposition of the cross-covariance matrix. We show that, although the model is underidentified, useful information may be extracted. We further consider an alternative parameterization in which one latent variable is associated with each block. Our analysis leads to some novel covariance equivalence results for Gaussian hidden variable models.

## 1 INTRODUCTION

Cross-covariance problems arise in the analysis of multivariate data that can be divided naturally into two blocks of variables, $\mathbf{X}$ and $\mathbf{Y}$, observed on the same units. In a cross-covariance problem we are interested, not in the within-block covariances, but in the way the $\mathbf{Y}$'s vary with the $\mathbf{X}$'s.

The field of behavioral teratology furnishes an example of a cross-covariance problem. In a study of the relationship between fetal alcohol exposure and neurobehavioral deficits reported by Sampson et al. [8] and by Streissguth et al. [11], $\mathbf{X}$ has thirteen columns, each corresponding to a different measure of the mother's reported alcohol consumption during pregnancy. $\mathbf{Y}$ has eleven columns, each corresponding to a different IQ subtest. The researchers are not primarily interested in the relationships between the different measures of the mother's alcohol intake or in the relationships between the different IQ subtests. They are interested in the relationship between alcohol intake and IQ. Neither of these phenomena can be measured directly.

A natural model to associate with the cross-covariance problem is the symmetric **paired latent correlation model**.[1] A path diagram (corresponding to a **semi-Markovian system of equations**, Pearl [6], pp. 30, 141) is seen in Figure 1. A formal specification of this paired latent model may be found in Section 2.2. With each block of observed variables is associated a latent variable, $\xi$ for the $\mathbf{X}$ block and $\omega$ for the $\mathbf{Y}$ block. The observed variables are linear functions of their parents, the latent variables, plus error. Correlated errors are indicated by bidirected edges. Under this model $\mathbf{X}$ and $\mathbf{Y}$ are conditionally independent given either or both of the latent variables.

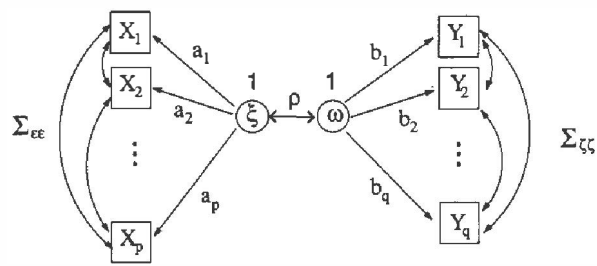

Figure 1: A symmetric paired latent correlation model.

A common approach in the factor analysis literature is to assume that within-block errors are uncorrelated.

---

[1]The term "symmetric" refers to the fact that both the $\mathbf{X}$'s and $\mathbf{Y}$'s are children of their respective latent variables, and that the latents have correlated errors. Asymmetric models will not be considered until Section 3.

This approach is incompatible, however, with our wish to model only between-block covariance. Thus in the current model the correlations of the within-block errors are unconstrained.

One problem with the latent model is that it is underidentified. That is, there are parameter values which cannot be distinguished on the basis of data. Furthermore it is not clear to which set of distributions over the observed variables the model corresponds. In this paper we overcome these problems by showing that the latent model corresponds to the set of all distributions over the observed variables in which the cross-covariance, $\Sigma_{XY}$, is of rank one. Consequently the latent model is appropriate for the setting we described, where we do not seek to model, and place no constraints on, the within-block covariance. Further this solution furnishes a precise answer to the question of identifiability. In estimating these models we are able to exploit well-developed methods: moment-based approaches using the singular value decomposition, and likelihood-based approaches used for reduced-rank regression.

As a corollary we prove covariance equivalence (see Pearl [6] p. 145) of two latent-variable models containing different numbers of latent variables. To our knowledge this is the first result of this kind. Note furthermore that we have specified this model to be symmetric in $X$ and $Y$. In fact it will turn out that there are asymmetric variants that are equivalent to the symmetric model. Finally we contrast model equivalence results in the case where the errors are unrestricted with the situation where they are assumed to be diagonal.

For related work which attempts to characterize the set of distributions over the observed variables induced by a latent model see Settimi and Smith [10] [9] and Geiger et al. [3].

## 2 MODELS

We introduce basic terms used to describe our result.

### 2.1 RANK-ONE CONSTRAINT MODELS

A rank-one **constraint model** is the set of $(p+q) \times (p+q)$ positive semidefinite matrices satisfying a rank constraint on the cross-covariance matrix:

$$\left. \begin{array}{c} \Sigma = \begin{bmatrix} \Sigma_{XX} & \Sigma_{XY} \\ \Sigma_{YX} & \Sigma_{YY} \end{bmatrix} , \\ \\ \mathrm{rank}(\Sigma_{XY}) = 1 , \quad \text{where } \Sigma_{XY} \text{ is } p \times q . \end{array} \right\} \quad (1)$$

### 2.2 PAIRED LATENT MODELS

The symmetric **paired latent correlation model** corresponds to the path diagram in Figure 1. The model is the set of distributions over the latent variables $\xi$ and $\omega$, the observed variables $X$ and $Y$, and the errors $\epsilon$ and $\zeta$, specified as follows.

$$\mathbf{Var}\begin{bmatrix} \xi \\ \omega \end{bmatrix} = \begin{bmatrix} 1 & \rho \\ \rho & 1 \end{bmatrix} ,$$

$$\mathbf{Var}(\epsilon) = \Sigma_{\epsilon\epsilon} ,$$

$$\mathbf{Var}(\zeta) = \Sigma_{\zeta\zeta} ,$$

$$\epsilon \perp\!\!\!\perp \begin{bmatrix} \xi \\ \omega \end{bmatrix} , \quad \epsilon \perp\!\!\!\perp \zeta , \quad \begin{bmatrix} \xi \\ \omega \end{bmatrix} \perp\!\!\!\perp \zeta ,$$

$$\mathbf{a} \in \mathbb{R}^p , \quad \mathbf{b} \in \mathbb{R}^q ,$$

$$\mathbf{x} = \mathbf{a}\xi + \epsilon , \quad \mathbf{y} = \mathbf{b}\omega + \zeta .$$

Thus the parameters are $\rho$, $\mathbf{a}$, $\mathbf{b}$, $\Sigma_{\epsilon\epsilon}$, and $\Sigma_{\zeta\zeta}$, where $|\rho| \leq 1$ and where $\Sigma_{\epsilon\epsilon}$, and $\Sigma_{\zeta\zeta}$ must be positive semidefinite. The vectors $\mathbf{a}$ and $\mathbf{b}$ are called **saliences** or "loadings."

The reader will observe that this model is underidentified. We shall precisely characterize the degree of non-identifiability, however, and suggest a natural convention which makes the model identifiable.

### 2.3 SINGLE LATENT MODELS

A symmetric **single latent model** is equivalent to a symmetric paired latent model where $\xi \equiv \omega$. See Figure 2. This model is the set of distributions over the latent variable $\eta$, the errors $\epsilon$ and $\zeta$, and the observed variables **X** and **Y**, specified as follows.

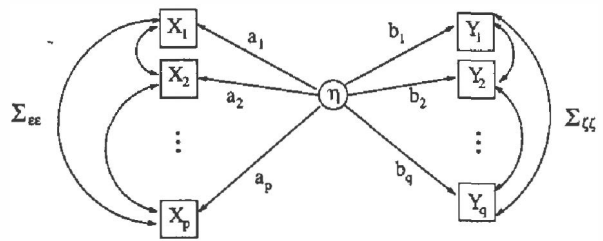

Figure 2: A symmetric single latent model.

$$\begin{aligned}
\text{Var}(\eta) &= 1, \\
\text{Var}(\epsilon) &= \Sigma_{\epsilon\epsilon}, \quad p \times p, \\
\text{Var}(\zeta) &= \Sigma_{\zeta\zeta}, \quad q \times q,
\end{aligned}$$

$$\epsilon \perp\!\!\!\perp \eta, \quad \epsilon \perp\!\!\!\perp \zeta, \quad \eta \perp\!\!\!\perp \zeta,$$

$$\mathbf{a} \in \mathbb{R}^p, \quad \mathbf{b} \in \mathbb{R}^q,$$

$$\mathbf{x} = \mathbf{a}\xi + \epsilon,$$

$$\mathbf{y} = \mathbf{b}\omega + \zeta.$$

Thus the parameters of the symmetric single latent model are $\Sigma_{\epsilon\epsilon}$, $\Sigma_{\zeta\zeta}$, **a** and **b**, where $\Sigma_{\epsilon\epsilon}$ and $\Sigma_{\zeta\zeta}$ must be positive semidefinite.

### 2.4 MAPS BETWEEN MODEL SPACES

The symmetric paired latent model induces a set of distributions over the observed variables as follows.

$$\left.\begin{aligned}
\Sigma_{XX} &= \mathbf{a}\mathbf{a}^T + \Sigma_{\epsilon\epsilon}, \\
\Sigma_{YY} &= \mathbf{b}\mathbf{b}^T + \Sigma_{\zeta\zeta}, \\
\Sigma_{XY} &= \mathbf{a}\mathbf{b}\rho.
\end{aligned}\right\} \quad (2)$$

The equations (2) define a map from the space of symmetric paired latent correlation models into the space of rank-one constraint models. The existence of such a map immediately raises the question whether every distribution in the rank-one constraint model can be obtained by a set of parameter values in the latent model—i.e., is the map onto. The answer is yes. The main result is stated as follows.

**Theorem 1** *For each distribution within the rank-one constraint model there is a non-void class of parameter values in the symmetric paired latent model which induce this distribution. We say that the paired latent model **parameterizes** the constraint model. Each constraint model distribution can in fact be parameterized by the single latent model.*

### 2.5 PROOF OF THE MAIN RESULT

We first show that any rank-one constraint model distribution can be parameterized by a single latent model. That is, suppose we are given a matrix $\Sigma$ satisfying (1). A set of parameter values satisfying

$$\left.\begin{aligned}
\Sigma_{XX} &= \mathbf{a}\mathbf{a}^T + \Sigma_{\epsilon\epsilon}, \\
\Sigma_{YY} &= \mathbf{b}\mathbf{b}^T + \Sigma_{\zeta\zeta}, \\
\Sigma_{\epsilon\epsilon} &\text{ is positive semidefinite}, \\
\Sigma_{\zeta\zeta} &\text{ is positive semidefinite},
\end{aligned}\right\} \quad (3)$$

$$\text{and } \Sigma_{XY} = \mathbf{a}\mathbf{b}^T \quad (4)$$

would parameterize the distribution. We shall show that it is always possible to find such parameter values. The proof uses two lemmas, stated in Section 4.

Decompose $\Sigma$ as follows:

$$\mathbf{w} = \begin{bmatrix} \mathbf{a} \\ \mathbf{b} \end{bmatrix}, \quad \mathbf{Q} = \mathbf{w}\mathbf{w}^T,$$

$$\mathbf{E} = \begin{bmatrix} \Sigma_{\epsilon\epsilon} & 0 \\ 0 & \Sigma_{\zeta\zeta} \end{bmatrix},$$

so that

$$\Sigma = \mathbf{Q} + \mathbf{E}.$$

Since $\Sigma_{XY}$ has rank one, by the singular value decomposition we can always find **a** and **b** satisfying (4). The two vectors are only determined up to sign and scale, however, since for any $\delta \neq 0$,

$$\Sigma_{XY} = \mathbf{a}\mathbf{b}^T \Rightarrow \Sigma_{XY} = (\delta\mathbf{a})\left(\frac{\mathbf{b}^T}{\delta}\right).$$

The scale and sign of **a** constitute the only degree of freedom, or lack of identifiability, in the map from the constraint model to the single latent model. This is because the direction of **a** is determined by (4). Once the sign and scale of **a** are determined, then **b** is determined by (4), and $\Sigma_{\epsilon\epsilon}$ and $\Sigma_{\zeta\zeta}$ are determined by (3).

Let us express the single degree of freedom in this model formally. Define **u** and **v** according to the convention of the singular value decomposition. That is, let

$$\Sigma_{XY} = \mathbf{u}\mathbf{v}^T d, \quad \|\mathbf{u}\| = \|\mathbf{v}\| = 1, \quad (5)$$

where $\|\cdot\|$ represents the Euclidean norm. Furthermore let us assume that a sign convention has been adopted, so that the lack of identifiability consists only in the scale of **a**. For $0 < \alpha$, let

$$\mathbf{a}(\alpha) \equiv \alpha\mathbf{u}, \quad \mathbf{b}(\alpha) \equiv \frac{\mathbf{v}d}{\alpha}. \quad (6)$$

Thus $\mathbf{a}(\alpha)$ and $\mathbf{b}(\alpha)$ satisfy $\Sigma_{XY} = \mathbf{a}(\alpha)[\mathbf{b}(\alpha)]^T$. To show that a latent parameterization exists it suffices to show that, if $\Sigma$ is positive semidefinite, a value of $\alpha$ can always be found such that the values determined by

$$\begin{aligned}
\Sigma_{\epsilon\epsilon}(\alpha) &\equiv \Sigma_{XX} - \mathbf{a}(\alpha)[\mathbf{a}(\alpha)]^T \\
\Sigma_{\zeta\zeta}(\alpha) &\equiv \Sigma_{YY} - \mathbf{b}(\alpha)[\mathbf{b}(\alpha)]^T
\end{aligned} \quad (7)$$

are positive semidefinite. Define $f : (0, \infty) \mapsto \mathbb{R}$ and $g : (0, \infty) \mapsto \mathbb{R}$ by

$$\begin{aligned}
f(\alpha) &= \min\{\text{eigenvalues of } \Sigma_{\epsilon\epsilon}(\alpha)\}, \\
g(\alpha) &= \min\{\text{eigenvalues of } \Sigma_{\zeta\zeta}(\alpha)\}.
\end{aligned} \quad (8)$$

It may be shown that these functions are continuous (Horn and Johnson [5]). Furthermore

- $f$ is monotone nonincreasing and goes to $-\infty$ as $\alpha \to \infty$;

- $g$ is monotone nondecreasing and goes to $-\infty$ as $\alpha \downarrow 0$

(see parts 1 and 3 of Lemma 6). Let

$$\mathcal{F} = \{\alpha : f(\alpha) < 0\}, \quad \text{and}$$
$$\mathcal{G} = \{\alpha : g(\alpha) < 0\}.$$

By the continuity of $f$ and $g$ these sets are open, but by monotonicity they are in fact intervals:

$$\mathcal{F} = (\alpha_1, \infty) \quad \text{and}$$
$$\mathcal{G} = (0, \alpha_2).$$

The closed set $\mathbb{R} \setminus (\mathcal{F} \cup \mathcal{G})$ is the set of feasible $\alpha$ values. By Lemma 7, this set is nonvoid; that is, we must have $\alpha_2 \leq \alpha_1$. Since this is the case, let us call them respectively $\alpha_{\min}$ and $\alpha_{\max}$. The feasible set of values for $\alpha$ is $[\alpha_{\min}, \alpha_{\max}]$, and we note the following:

$$\alpha_{\min} = \min\{\alpha : \Sigma_{\zeta\zeta}(\alpha) \text{ is positive semidefinite}\},$$

$$\alpha_{\max} = \max\{\alpha : \Sigma_{\epsilon\epsilon}(\alpha) \text{ is positive semidefinite}\}. \quad (9)$$

These follow from the definitions at (7) and (8).

The fact that $\mathbb{R} \setminus (\mathcal{F} \cup \mathcal{G})$ is nonvoid means the following: In equations (3) and (4) there is at least one scale of the salience vector $\mathbf{a}$ such that both $\Sigma_{\epsilon\epsilon}$ and $\Sigma_{\zeta\zeta}$ are positive semidefinite. Thus there is a single-latent parameterization of any distribution in the rank-one constraint model. If $\alpha_{\min} = \alpha_{\max}$, there is only one parameterization, say $\alpha^*$. Since $f$ and $g$ are continuous, $f(\alpha^*) = g(\alpha^*) = 0$ and the unique parameterization yields singular within-block error variances for both blocks. If $\alpha_{\min} < \alpha_{\max}$ nonsingular parameterizations will usually exist. Examples exist, however, where $\alpha_{\min} < \alpha_{\max}$ and all parameterizations are singular; see Wegelin et al. [12].

### 2.6 PAIRED LATENT MODEL

A distribution in the rank-one constraint model may be mapped to an equivalence class of parameter values in the symmetric paired latent correlation model as follows. We define the feasible set of values for $\alpha$ and $\rho$ to be

$$\{(\rho, \alpha) : \quad |\rho| \leq 1, \quad \alpha_{\min}^2 \leq \alpha^2 \rho^2, \quad \alpha \leq \alpha_{\max}\}.$$

When $\alpha_{\min} = \alpha_{\max}$ this set is a singleton; otherwise it is a continuous closed region. An example may be seen in Figure 3. Each feasible point $(\rho, \alpha)$ determines a complete set of parameter values in the symmetric paired latent correlation model as follows:

$$\mathbf{a} = \alpha \mathbf{u},$$
$$\mathbf{b} = \frac{\mathbf{v}}{\alpha \rho} d,$$
$$\Sigma_{\epsilon\epsilon} = \Sigma_{XX} - \mathbf{a}\mathbf{a}^T,$$
$$\Sigma_{\zeta\zeta} = \Sigma_{YY} - \mathbf{b}\mathbf{b}^T.$$

Note that $\mathbf{Cor}(\xi, \omega) = 1$ is always feasible. Observing that

$$\alpha^2 \leq \alpha_{\max}^2 \quad \text{and} \quad \alpha_{\min}^2 \leq \rho^2 \alpha^2 \quad \Rightarrow \quad |\rho| \geq \frac{\alpha_{\min}}{\alpha_{\max}}$$

we see that the correlation is bounded below, and we define

$$\rho_{\min} = \frac{\alpha_{\min}}{\alpha_{\max}}.$$

Similarly to the previous case, values outside the feasible set would lead to at least one of the three covariance matrices failing to be positive semidefinite.

### 2.7 EXAMPLE

Consider the following nonsingular matrix.

$$\Sigma = \begin{bmatrix} 7 & 0 & 0 & 1 & 0.5 \\ 0 & 7 & 0 & 2 & 1 \\ 0 & 0 & 7 & 3 & 1.5 \\ 1 & 2 & 3 & 9 & 0 \\ 0.5 & 1 & 1.5 & 0 & 5 \end{bmatrix} \quad (10)$$

Let $p = 3$, $q = 2$. Choosing the convention that both $d$ and the component of $\mathbf{u}$ with greatest absolute value shall be positive, we obtain

$$\mathbf{a}(\alpha) = \frac{\alpha}{\sqrt{14}} \begin{bmatrix} 1 \\ 2 \\ 3 \end{bmatrix}, \quad \mathbf{b}(\alpha) = \frac{\sqrt{14}}{2\alpha} \begin{bmatrix} 2 \\ 1 \end{bmatrix},$$

$$d = \sqrt{\frac{35}{2}},$$

$$\det \Sigma_{\epsilon\epsilon}(\alpha) = 343 - 49\alpha^2,$$
$$\det \Sigma_{\zeta\zeta}(\alpha) = 45 - \frac{203}{2\alpha^2},$$

$$[\alpha_{\min}, \alpha_{\max}] = \left[\frac{1}{30}\sqrt{2030}, \sqrt{7}\right]$$

$$\approx [1.50, 2.65].$$

Thus $\rho_{\min} \equiv \frac{\alpha_{\min}}{\alpha_{\max}} = \frac{1}{30}\sqrt{290} \approx 0.57$. The feasible set is displayed in Figure 3.

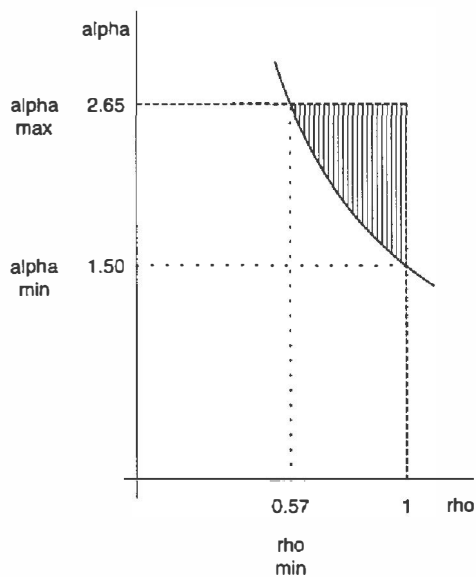

Figure 3: Feasible $\rho$ and $\alpha$ for the paired-latent correlation parameterization of the rank-constraint distribution specified by (10). Correlation $\rho$ is on the x-axis, $\alpha$ on the y-axis. Feasible values are in the shaded region in the upper right corner of the rectangle. The right boundary of the feasible set corresponds to the single latent model.

### 2.8 DISCUSSION

Three spaces of covariance matrices over the observed variables **X** and **Y** are of interest in the current work. They are:

1. Those corresponding to the constraint model.
2. Those induced by the symmetric paired latent correlation model.
3. Those induced by the symmetric single latent model.

It follows from definitions and from Equations (2) that Set 3 $\subset$ Set 2 $\subset$ Set 1. Theorem 1, however, implies that Set 1 $\subset$ Set 3. Hence Set 1 = Set 2 = Set 3, a fact which we state as the following corollary.

**Corollary 2** *The sets of covariance matrices over the observed variables induced by the symmetric paired latent correlation model and the symmetric single latent model are equal to the set of covariance matrices belonging to the rank-one constraint model.*

Thus there is no way using only data to distinguish between the three models. Furthermore it is well-known that the rank-one constraint model is covariance equivalent to reduced-rank regression (RRR). Since maximum-likelihood estimation procedures are available for RRR (Anderson [1]), the problems of maximum-likelihood estimation for the paired and single symmetric latent models are solved.

The choice of the quantity $\rho$ within the feasible interval $[\rho_{\min}, 1]$ entails a tradeoff. When $\rho = \rho_{\min}$ the error variances for both blocks are at their minimum, and in fact the covariance matrices $\Sigma_{\epsilon\epsilon}$ and $\Sigma_{\zeta\zeta}$ are singular. When $\rho = 1$, on the other hand, at least one of the error variances may be nonsingular. Thus when $\rho_{\min} < 1$ we choose either to have latent variables perfectly correlated but poorly measured, or latent variables measured with minimal error, but poorly correlated with each other.

For any fixed $\rho > \rho_{\min}$ there are many feasible values for $\alpha$. Greater values of $\alpha$ correspond to error variance which is less for the **X** block and greater for for the **Y** block.

For other applications of the singular value decomposition in AI, see Dumais [2].

## 3 RELATED EQUIVALENCE RESULTS

In this section we extend the results described so far by considering a number of other latent models which relate the two blocks of observed variables.

These graphs are shown in Figure 4. (a) and (b) represent two path diagrams in which the latent variables $\xi$ and $\omega$ are parents of the observed variables. The only difference between the models is that (a) specifies that $\xi$ and $\omega$ are correlated, while in (b) $\xi$ is a parent of $\omega$. The graph shown in Figure 4 (c) differs from that shown in (b) in that the **X** variables are parents of $\xi$. The graph in (d) is analogous to (a) and (b) but the pair of latent variables $\xi, \omega$ are replaced with a single variable. Likewise (e) represents the single latent analogue to (c).

We consider the five models corresponding to these graphs, under two sets of conditions on the error terms:

(I) $\text{Cov}(\epsilon_i, \zeta_j) = 0$, but $\text{Cov}(\epsilon_i, \epsilon_k)$ and $\text{Cov}(\zeta_j, \zeta_\ell)$ are unrestricted.

(II) $\text{Cov}(\epsilon_i, \zeta_j) = 0$, and

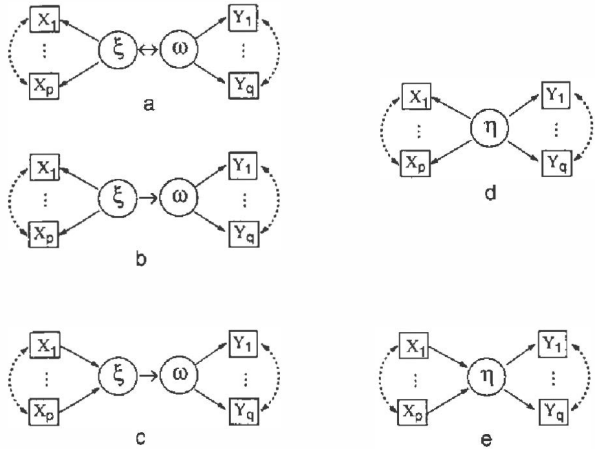

Figure 4: Path diagrams corresponding to two-block latent variable models. Under (I) the dashed edges are present; under (II) they are absent. Under (I) all models are covariance equivalent over **X** and **Y**.

$$\mathbf{Cov}\left(\epsilon_i, \epsilon_k\right) = \mathbf{Cov}\left(\zeta_j, \zeta_\ell\right) = 0 \text{ for } i \neq k,\, j \neq \ell.$$

Let $\mathcal{N}_a^{\mathrm{I}}$ denote the set of Gaussian distributions over **X** and **Y** given by graph (a) in Figure 4 under condition (I) on the errors, likewise for $\mathcal{N}_a^{\mathrm{II}}$, $\mathcal{N}_b^{\mathrm{I}}$, $\mathcal{N}_b^{\mathrm{II}}$ and so on. Corollary 2 of the previous section thus shows that $\mathcal{N}_a^{\mathrm{I}} = \mathcal{N}_d^{\mathrm{I}}$. We extend these results further in the next theorem.

**Theorem 3** *The following relations hold:*

$$\mathcal{N}_a^{\mathrm{I}} = \mathcal{N}_b^{\mathrm{I}} = \mathcal{N}_c^{\mathrm{I}} = \mathcal{N}_d^{\mathrm{I}} = \mathcal{N}_e^{\mathrm{I}}$$

$$\mathcal{N}_a^{\mathrm{II}} = \mathcal{N}_b^{\mathrm{II}} \neq \mathcal{N}_c^{\mathrm{II}} = \mathcal{N}_e^{\mathrm{II}} \neq \mathcal{N}_d^{\mathrm{II}} \neq \mathcal{N}_a^{\mathrm{II}}$$

*(The first and third inequalities require $p > 1$. The second also requires $q > 1$.)*

In words: When the within-block errors are not restricted, all of the latent structures in Figure 4 are indistinguishable. When the errors are uncorrelated, on the other hand, the following conditions hold:

- We can distinguish structures in which $\xi$ is a parent of the **X**'s from those in which the **X**'s are parents of $\xi$.

- When the **X**'s are parents of $\xi$ we cannot distinguish between models with one and two latent variables.

- When $\xi$ is a parent of the **X**'s we can distinguish models with two latent variables from those containing only one.

The existence of equivalent models containing different numbers of hidden variables is important for the purpose of interpretation. It highlights the danger of postulating the existence of variables for which there is no evidence in the data.

### 3.1　PROOFS OF EQUIVALENCE RESULTS

In order to prove the results in Theorem 3 we need several definitions. Following [7] we say that a path diagram, which may contain directed edges ($\rightarrow$) and bi-directed edges ($\leftrightarrow$) is *ancestral* if:

(a) there are no directed cycles;

(b) if there is an edge $x \leftrightarrow y$ then $x$ is not an ancestor of $y$, (and vice versa);

where a vertex $x$ is said to be an *ancestor* of $y$ if either $x = y$ or there is a directed path from $x$ to $y$. Conditions (a) and (b) may be summarized by saying that if $x$ and $y$ are joined by an edge and there is an arrowhead at $x$ then $x$ is *not* an ancestor of $y$; this is the motivation for the term 'ancestral'. (In [7] a more general version of this definition is given which applies to graphs containing undirected edges.)

A natural extension of Pearl's d-separation criterion may be applied to graphs containing directed and bi-directed edges. A non-endpoint vertex $v$ on a path is said to be a *collider* if two arrowheads meet at $v$, i.e. $\rightarrow v \leftarrow$, $\leftrightarrow v \leftrightarrow$, $\leftrightarrow v \leftarrow$ or $\rightarrow v \leftrightarrow$; all other non-endpoint vertices on a path are *non-colliders*. A path $\pi$ between $\alpha$ and $\beta$ is said to be *m-connecting given* $Z$ if the following hold:

(i) no non-collider on $\pi$ is on $Z$;

(ii) every collider on $\pi$ is an ancestor of a vertex in $Z$.

Two vertices $\alpha$ and $\beta$ are said to be m-separated given $Z$ if there is no path m-connecting $\alpha$ and $\beta$ given $Z$. Disjoint sets of vertices $A$ and $B$ are said to be m-separated given $Z$ if there is no pair $\alpha, \beta$ with $\alpha \in A$ and $\beta \in B$ such that $\alpha$ and $\beta$ are m-connected given $Z$. (This an extension of the original definition of d-separation for DAGs in that the notions of 'collider' and 'non-collider' now include bi-directed edges.) Two graph $\mathcal{G}_1$ and $\mathcal{G}_2$ are said to be *Markov equivalent* if for all disjoint sets $A, B, Z$ (where $Z$ may be empty), $A$ and $B$ are m-separated given $Z$ in $\mathcal{G}_1$ if and only if $A$ and $B$ are m-separated given $Z$ in $\mathcal{G}_2$. A distribution $P$ is said to obey the *global Markov property with respect to graph* $\mathcal{G}$ if $A \perp\!\!\!\perp B \mid Z$ in $P$ whenever $A$ is m-separated from $B$ given $Z$ in $\mathcal{G}$.

An ancestral graph is said to be *maximal* if for every pair of non-adjacent vertices $\alpha, \beta$ there exists some set $Z$ such that $\alpha$ and $\beta$ are m-separated given $Z$.

It is proved in [7] that the set of Gaussian distributions given by parameterizing the path diagram $\mathcal{G}$ is exactly the set of Gaussian distributions that obey the global Markov property with respect to $\mathcal{G}$. More formally, we have:

**Theorem 4** *If $\mathcal{G}$ is a maximal ancestral graph then the following equality holds regarding Gaussian distributions:*

$\{N \mid N$ results from some assignment of parameter values to $\mathcal{G}\}$

$= \{N \mid N$ satisfies the global Markov property for $\mathcal{G}\}$.

See Theorem 8.14 in [7]. As an immediate Corollary we have:

**Corollary 5** *If $\mathcal{G}_1$ and $\mathcal{G}_2$ are two Markov equivalent maximal ancestral graphs then they parameterize the same sets of Gaussian distributions.*

See Corollary 8.19 in [7]. These results do not generally hold for path diagrams which are not both maximal and ancestral.

The sets of distributions given by the models under (I) correspond to the path diagrams shown in Figure 4 in which there are bi-directed edges between all variables within the same block, thus $\mathbf{X}_i \leftrightarrow \mathbf{X}_k$ ($i \neq k$) and $\mathbf{Y}_j \leftrightarrow \mathbf{Y}_\ell$ ($j \neq \ell$).

### 3.2 PROOF OF THEOREM 3

We first show $\mathcal{N}_a^I = \mathcal{N}_b^I = \mathcal{N}_c^I$. Observe that in each of the graphs in Figure 4(a), (b) and (c), the following m-separation relations hold:

(i) $\mathbf{X}_i$ is m-separated from $\mathbf{Y}_j$ by any non-empty subset of $\{\boldsymbol{\xi}, \boldsymbol{\omega}\}$;

(ii) $\mathbf{X}_i$ is m-separated from $\boldsymbol{\omega}$ by $\boldsymbol{\xi}$;

(iii) $\mathbf{Y}_j$ is m-separated from $\boldsymbol{\xi}$ by $\boldsymbol{\omega}$.

Further, when bi-directed edges are present between vertices within each block all other pairs of vertices are adjacent so there are no other m-separation relations. Consequently these graphs are Markov equivalent and maximal since there is a separating set for each pair of non-adjacent vertices. It then follows directly by Corollary 5 that these graphs parameterize the same sets of distributions over the set $\{\mathbf{X}, \mathbf{Y}, \boldsymbol{\omega}, \boldsymbol{\xi}\}$, hence they induce the same sets of distributions on the margin over $\{\mathbf{X}, \mathbf{Y}\}$.

The proof that $\mathcal{N}_d^I = \mathcal{N}_e^I$ is very similar. When bi-directed edges are present within each block the only pairs of non-adjacent vertices are $\mathbf{X}_i$ and $\mathbf{Y}_j$ which are m-separated by $\boldsymbol{\xi}$. It then follows as before that these graphs are Markov equivalent and maximal and hence by Corollary 5 they parameterize the same sets of distributions over $\{\mathbf{X}, \mathbf{Y}, \boldsymbol{\xi}\}$, and consequently over $\{\mathbf{X}, \mathbf{Y}\}$.

Since we have already shown $\mathcal{N}_a^I = \mathcal{N}_d^I$ in Corollary 2, the proof of equivalences concerning models with error structure given by (I) is complete. It remains to prove the results concerning models of type (II). These correspond to the path diagrams in Figure 4, without the dashed edges between vertices within the same block. Subsequent references to graphs in this figure will be to the graphs without these within-block edges.

First note that the m-separation relations given by (i), (ii), (iii) above continue to hold when there are no edges between vertices within each block. In graphs (a) and (b) we also have:

(iv) $\mathbf{X}_i$ and $\mathbf{X}_j$ are m-separated given $\boldsymbol{\xi}$;

(v) $\mathbf{Y}_i$ and $\mathbf{Y}_j$ are m-separated given $\boldsymbol{\omega}$.

Consequently these graphs are Markov equivalent and maximal. Hence $\mathcal{N}_a^{II} = \mathcal{N}_b^{II}$ by Corollary 5. In the path diagrams corresponding to (c) and (e), we have

(vi) $\mathbf{X}_i$ and $\mathbf{X}_j$ are m-separated by the empty set.

Consequently the variables in the $\mathbf{X}$ block are marginally independent in $\mathcal{N}_c^{II} = \mathcal{N}_e^{II}$, while this is not so under $\mathcal{N}_a^{II}, \mathcal{N}_b^{II}, \mathcal{N}_d^{II}$. This establishes two of the inequalities. By direct calculation it may be seen that for any distribution in $\mathcal{N}_d^{II}$ it holds that

$$\mathrm{Cov}\,(\mathbf{X}_i, \mathbf{X}_k)\,\mathrm{Cov}\,(\mathbf{Y}_j, \mathbf{Y}_\ell) = \mathrm{Cov}\,(\mathbf{X}_i, \mathbf{Y}_j)\,\mathrm{Cov}\,(\mathbf{X}_k, \mathbf{Y}_\ell)$$

while this does not hold for distributions in $\mathcal{N}_a^{II} = \mathcal{N}_b^{II}$. This establishes the third inequality. It only remains to show that $\mathcal{N}_c^{II} = \mathcal{N}_e^{II}$. First observe that the set of m-separation relations which hold among $\{\mathbf{X}, \mathbf{Y}, \boldsymbol{\omega}\}$ in the graph in (c), i.e. (i), (ii), (iii) and (v), is identical to the set of relations holding among $\{\mathbf{X}, \mathbf{Y}, \boldsymbol{\eta}\}$ in (e), i.e. (i), (ii), (iii) and

(vii) $\mathbf{Y}_j$ and $\mathbf{Y}_\ell$ are m-separated by $\boldsymbol{\eta}$,

where $\eta$ is substituted for $\omega$. Consequently any marginal distribution over $\{\mathbf{X}, \mathbf{Y}, \boldsymbol{\xi}\}$ that is obtained from the graph in (c) may also be parameterized by the graph in (e) after substituting $\eta$ for $\omega$. It then follows that $\mathcal{N}_c^{II} \subseteq \mathcal{N}_e^{II}$. To prove the opposite inclusion it is sufficient to observe that any distribution over $\{\mathbf{X}, \mathbf{Y}, \boldsymbol{\eta}\}$ that is parameterized by the graph in (e) may be parameterized by the graph in (c) by setting $\omega = \xi + \epsilon_\omega$ and letting $\mathbf{Var}(\epsilon_\omega) + \mathbf{Var}(\epsilon_\xi) = \mathbf{Var}(\epsilon_\eta)$. This completes the proof. □

## 4 APPENDIX

The following lemmas are proved in Wegelin et al. [12].

**Lemma 6** *Let* $\mathbf{A}$ *and* $\mathbf{C}$ *be symmetric matrices of the same dimension,* $\mathbf{C}$ *positive semidefinite. Let* $h : (0, \infty) \mapsto \mathbb{R}$ *be defined by*

$h(\alpha) = $ *the smallest eigenvalue of* $(\mathbf{A} - \alpha \mathbf{C})$ .

*Then*

1. *The function $h$ is monotone nonincreasing. If $\mathbf{C}$ is strictly positive definite, the function is strictly monotone decreasing.*

2. $\lim_{\alpha \downarrow 0} h(\alpha) = h(0)$.

3. *If $\mathbf{C}$ has at least one positive eigenvalue, $\lim_{\alpha \uparrow \infty} h(\alpha) = -\infty$.*

**Lemma 7** *Let*

$$\boldsymbol{\Sigma} = \left[ \begin{array}{cc} \mathbf{A} & \mathbf{C} \\ \mathbf{C}^T & \mathbf{B} \end{array} \right],$$

*where $\boldsymbol{\Sigma}$ is symmetric positive semidefinite, $\mathbf{A}$ and $\mathbf{B}$ are respectively $p \times p$ and $q \times q$, and $\mathbf{C}$ is of rank one. Let $\mathbf{u}$ and $\mathbf{v}$ be $p$- and $q$-vectors satisfying $\mathbf{C} = \mathbf{u}\mathbf{v}^T$. Define $\mathbf{A}^* = \mathbf{A} - \mathbf{u}\mathbf{u}^T$, $\mathbf{B}^* = \mathbf{B} - \mathbf{v}\mathbf{v}^T$. Then at least one of $\mathbf{A}^*$ and $\mathbf{B}^*$ is positive semidefinite. Furthermore, if $\boldsymbol{\Sigma}$ is positive definite, at least one of $\mathbf{A}^*$ and $\mathbf{B}^*$ is positive definite.*

### Acknowledgements

This research was supported in part by National Science Foundation Grants Nos. DMS-9972008 and DMS-0071818.